\documentclass{pf2007}

\usepackage[frenchb]{babel}
\usepackage[utf8]{inputenc}
\usepackage[T1]{fontenc}
\usepackage[usenames]{color}
\usepackage{natbib,xspace,amsmath,amsfonts,amssymb,rotating,url,listings,caption}

\usepackage{float}
\floatstyle{boxed} 
\restylefloat{figure}

\def\taaable{{\sc Taaable}\xspace}

\shorttitle{Formalismes temporels pour les recettes}
\shortouvrage{RTE 2011}

\title{Quels formalismes temporels pour\\représenter des connaissances extraites de\\textes de recettes de cuisine ?}
\author{Valmi Dufour-Lussier\inst{1,}\inst{2} \and Florence Le Ber\inst{1,}\inst{3} \and Jean Lieber\inst{1,}\inst{4}}
\institute{
	LORIA (CNRS, INPL, INRIA, UN2, UHP)\\
	615, rue du Jardin Botanique, 54602 Villers-lès-Nancy\\
	valmi.dufour@loria.fr, jean.lieber@loria.fr
 \and
   Université Nancy 2
 \and
 	LHYGES (CNRS, ENGEES, Université de Strasbourg)\\
 	1, quai Koch, 67000 Strasbourg\\
 	florence.leber@engees.unistra.fr
 \and
   Université Henri Poincaré, Nancy 1
}

\begin{document}
\maketitle

\begin{abstract}
Le projet \taaable a pour objet de construire un syst\`eme de
raisonnement \`a partir de cas pour la recherche et l'adaptation de
textes de recettes de cuisine. Dans ce cadre, nous nous int\'eressons aux aspects
temporels des recettes et \`a la mani\`ere de les repr\'esenter afin d'en
envisager l'adaptation textuelle.

  \motscles{Adaptation, algèbres temporelles, traitement automatique des langues}
\end{abstract}

\section{Introduction}

\taaable \citep{taaable08} est un système informatique destiné à résoudre des problèmes de cuisine, développé dans le cadre du \textit{Computer Cooking Contest\footnote{\url{http://liris.cnrs.fr/ccc/ccc2010}},} dans le cadre duquel des requêtes du type \og je désire la recette d'un plat de pâtes aux lentilles \fg doivent être résolues en cherchant dans un livre fourni par les organisateurs du concours une recette qui répond aux critères de l'utilisateur. Si une telle recette n'existe pas, un moteur de raisonnement à partir de cas sélectionne une recette similaire et suggère une adaptation. Initialement, l'adaptation ne portait que sur la liste d'ingrédients, mais la dernière version du système \citep{taaable10} tient compte des instructions textuelles de préparation dans le raisonnement et l'adaptation. Le raisonnement ne pouvant porter directement sur du texte brut, il est nécessaire de choisir un formalisme de représentation de cas et d'employer des méthodes de traitement automatique des langues pour créer ces représentations. \taaable utilise actuellement un formalisme \textit{ad hoc} fondé sur les combinaisons d'ingrédients pour ce faire, bien que \cite{lln09rapc,lln10rte} aient initié une réflexion quant à un formalisme plus expressif et plus approprié.

Cet article poursuit cette réflexion. Dans la section~\ref{sec:phen}, nous présentons les différents types de phénomènes temporels observés dans les textes de recette, des plus simples à représenter aux plus complexes. Puis, dans la section~\ref{sec:rais}, nous définissons le type de raisonnement que nous entendons appliquer aux recettes avant de passer en revue, dans la section~\ref{sec:form}, quelques formalismes qui semblent appropriés à la représentation de ces phénomènes. La section~\ref{sec:synth} présente de façon concise l'adéquation que nous observons entre les phénomènes à modéliser et les formalismes étudiés. Finalement, dans la section~\ref{sec:concl}, nous concluons et présentons succinctement les travaux futurs que nous comptons accomplir à partir de nos résultats.

\section{Problématiques de représentation temporelle}
\label{sec:phen}

Nous dressons d'abord une liste des phénomènes temporels que le formalisme retenu devrait être capable de représenter. La figure~\ref{fig:recette} présente une recette du livre employé par \taaable pour répondre aux requêtes des utilisateurs, reproduite telle quelle.

\begin{figure}
	\textbf{Lutheran Hotdish}
	\begin{itemize}
		\item 1 lb Hamburger
		\item 1/2 lb Mild or spicy sausage
		\item 1 lg Onion (sliced and quartered) (up to)
		\item 5 Cloves garlic (minced) (up to)
		\item 2 cn Kidney beans (drained)
		\item 1 lb Uncooked pasta (i.e. elbow; twisted; wagon wheels, shells; etc) (up to)
	\end{itemize}
	Brown hamburger and sausage with onion, garlic, and all seasonings. Meanwhile, prepare the pasta per pkg instructions. In a large pan, combine all ingredients. Add enough tomato sause until mixture is well coated, but no soupy. When well mixed, pour into greased and covered casserol dish and bake at 350F for 1hr. Remove cover for the last 15 minutes. Then enjoy.
	\caption{Une recette issue du livre fourni par le \textit{Computer Cooking Contest,} illustrant certains phénomènes temporels intéressants.}
	\label{fig:recette}
\end{figure}

\subsection{Durée}
\label{subsec:duree}

Pour les événements ayant une durée, celle-ci peut être définie explicitement de manière quantitative, comme c'est le cas des dernières instructions de la recette montrée en figure~\ref{fig:recette} (\og bake at 350F for 1hr\fg), ou encore implicitement en référence à un changement d'état (\og Add enough tomato sauce until mixture is well coated\fg), ce que nous appellerons \og durée qualitative \fg. Certaines recettes peuvent préciser les deux (\og Bake at 350 degrees for about 25 minutes, or until lightly browned\fg), auquel cas une sémantique doit être déterminée : s'agit-il réellement d'une disjonction --- l'action se conclut-elle dès que 25 minutes se sont écoulées \emph{ou} que la couleur a changé ?

\subsection{Ordre}
\label{subsec:ordre}

Les actions nécessaires à l'exécution d'une recette de cuisine ne peuvent pas être effectuées dans n'importe quel ordre. La plupart des actions sont ordonnées implicitement, puisqu'il est entendu que, dans une recette de cuisine, les actions décrites à même la liste d'ingrédients sont à accomplir avant le début de la recette proprement dite, et que les actions décrites dans le texte doivent être effectuées dans l'ordre où elles sont énoncées, sauf mention contraire.

Dans la recette de la figure~\ref{fig:recette}, les actions préliminaires (couper l'oignon, hacher l'ail et égoutter les haricots) doivent être effectuées avant que la première action décrite dans le texte (\og Brown hamburger and sausage...\fg) soit initiée. Cependant il semble que ces actions puissent être exécutées dans n'importe quel ordre, puisqu'elles ne dépendent pas l'une des autres : on peut couper l'oignon avant de hacher l'ail ou après, car l'aliment résultant de chacune de ces actions n'est pas utilisé pour l'autre.

Dans le texte même, certains adverbes ont pour effet de changer l'ordre normal d'exécution des actions. Par exemple, l'action décrite par la deuxième phrase (\og Meanwhile, prepare pasta...\fg) doit être effectuée simultanément à la première action. La troisième phrase (\og In a large pan, combine all ingredients\fg) ne précise pas de relation d'ordre explicite, de quoi on peut conclure que l'action doit être effectuée à la suite de l'action précédente.

Il est à noter que le moment d'exécution d'une action peut être défini en référence à la fin d'un intervalle, comme dans \og Remove cover for the last 15 minutes.\fg Cet intervalle de référence peut être imprécis, comme dans \og simmer 2--3 hours \fg, ajoutant une difficulté additionnelle.

\subsection{Actions répétées}
\label{subsec:repet}

Les actions répétées suggèrent une structure non linéaire du temps. Différentes situations peuvent poser différents problèmes de représentation. Une même action peut être répétée un nombre déterminé ou indéterminé (jusqu'à l'obtention d'un état donné) de fois, ou une série d'actions peuvent être répétées en alternance, possiblement avec des conditions d'ordre précises. Une action peut également être répétée de façon sporadique sur un intervalle donné, comme dans \og simmer 2--3 hours, stirring occasionally \fg.

\subsection{Disjonction}
\label{subsec:disj}

Les disjonctions d'actions constituent un dernier élément pouvant poser des difficultés de représentation. Celles-ci peuvent être liées à un état, ou encore représenter des alternatives dans une recette : \og If you want this relish to be really hot then take 4 green chillis, seed them and chop finely, and put them in the frying pan at the same time as the onions. \fg

\section{Raisonnement}
\label{sec:rais}

Avant de discuter des formalismes que nous envisageons d'utiliser pour représenter les recettes, il convient de mieux définir le type de raisonnement que nous souhaitons appliquer à ces recettes. Si l'utilisateur lance la requête donnée en exemple dans l'introduction, c'est-à-dire un plat de pâtes aux lentilles, \taaable retournera comme résultat la recette de la figure~\ref{fig:recette} en suggérant de remplacer les haricots rouges en conserve par des lentilles.

Pour adapter le texte de préparation de manière adéquate, il ne suffira pas de remplacer toutes les instances d'\og haricots \fg par \og lentilles \fg. Il faut également introduire au bon endroit des informations issues des connaissances du domaine : par exemple, que les lentilles doivent cuire à l'eau \emph{pendant} une demi-heure \emph{puis} être égouttées \emph{avant} d'intégrer la préparation. Pour représenter adéquatement les relations entre ces actions (pendant, puis, avant), il sera nécessaire d'employer une représentation temporelle.

Une manière de tenir compte de connaissances liées à un élément de substitution en raisonnement à partir de cas est l'adaptation par révision, proposée par \cite{cl08iccbr} : il s'agit d'adjoindre ces connaissances à celles issues du cas remémoré (ici, la recette du \og Lutheran Hotdish \fg) en résolvant les contradictions éventuelles à l'aide d'un opérateur de révision des croyances, de manière à conserver toute l'information issue des connaissances du domaine et le maximum d'information issue du cas.

Un tel opérateur applicable aux algèbres qualitatives pourrait être dérivé des travaux de \cite{ckms09cosit} sur la fusion des réseaux de contraintes qualitatives.

Une fois la représentation de la préparation de la recette adaptée par révision avec les connaissances du domaine, et à condition de pouvoir maintenir le lien entre le texte et cette représentation au cours de l'adaptation, il sera possible d'apporter les modifications correspondantes au texte.

\section{Formalismes envisagés}
\label{sec:form}

Les formalismes de représentation temporelle sont généralement divisés en deux catégories par la communauté IA, en fonction du type de raisonnement envisagé. Les raisonnements sur les contraintes temporelles font généralement appel à des formalismes algébriques tels que les algèbres d'intervalle, que nous étudierons en particulier. Les raisonnements sur les actions et leurs effets sur des domaines dynamiques, quant à eux, donnent le plus souvent lieu à l'utilisation de formalismes inspirés du calcul des situations, qui sont moins appropriés à notre problème\footnote{
Nous n'entendons pas que ces formalismes ne seraient pas utiles pour l'adaptation dans \taaable, et au contraire nous souhaitons étudier la contribution qu'ils pourraient apporter ultérieurement, mais le problème que nous cherchons à résoudre ici est relatif aux contraintes temporelles entre les actions, et non pas au raisonnement sur les actions et leurs effets.
}.

Nous nous pencherons brièvement sur des formalismes dont le but principal est de résoudre certains problèmes de représentation des informations temporelles dans les textes, ainsi que sur les flux opérationnels \textit{(workflows),} représentant des processus et pouvant être utilisés pour raisonner sur ceux-ci.

\subsection{Algèbres de relations}

\subsubsection{Contraintes qualitatives}

Nous débutons notre exploration par l'algèbre des intervalles de \cite{allen83cacm}, qui nous semble particulièrement adaptée à la représentation des phénomènes les plus fréquents. Cet algèbre permet d'exprimer des relations entre des intervalles temporels sous la forme de disjonctions entre treize relations de base : \textit{before} (notée b, $<$ ou p), \textit{meets} (m), \textit{overlaps} (o), \textit{during} (d), \textit{starts} (s), \textit{finishes} (f), les inverses de ces six relations (respectivement bi --- ou encore $>$, pi ou a ---, mi, oi, di, si et fi), ainsi que \textit{equals} (e, eq ou $=$).

En associant chaque action à un intervalle, il est facile d'en représenter l'ordre. Dans la recette de la figure~\ref{fig:recette}, on aurait par exemple :
\begin{itemize}
	\item `mince garlic' \{b\} `brown hamburger'
	\item `prepare pasta' \{d,f\} `brown hamburger'
	\item `combine all ingredients' \{bi,m\} `prepare pasta'
\end{itemize}
En associant des états à des intervalles, on pourrait représenter les durées définies par un changement d'état, comme dans le cas de \og `add tomato sauce' \{m\} `mixture is well coated' \fg.

Grâce à $\mathcal{INDU}$ \citep{pks99atai}, qui permet additionnellement d'exprimer des contraintes qualitatives sur la durée des intervalles (par exemple, un intervalle $I_1$ dont la fin coïncide avec le début d'un intervalle $I_2$ et dont la durée est inférieure à celle de ce dernier s'exprime \og $I_1$ \{m$^\leqslant$\} $I_2$ \fg), il serait également possible de représenter jusqu'à un certain point des durées quantitatives en définissant des intervalles de durée fixe, par exemple : \og `bake' \{?$^=$\} `une heure' \fg. Cette même méthode nous permettrait de représenter aussi la dernière action de la recette : \og `remove cover' \{s$^?$\} `15 minutes' \{f$^?$\} `bake' \fg. Pour représenter la durée d'une qui est exprimée à la fois quantitativement et qualitativement, comme dans \og Bake at 350 degrees for about 25 minutes, or until lightly brown \fg, nous pouvons prendre le parti d'exprimer que la durée de 25 minutes constitue un maximum et écrire \og `bake' \{m$^?$\} `is brown', `bake' \{?$^\leqslant$\} `25 minutes' \fg.

\cite{ladkin86aaai} et \cite{ligozat91aaai}, entre autres, ont proposé des extensions à l'algèbre de Allen afin qu'elle puisse exprimer des contraintes sur des intervalles non convexes. En définissant une propriété \og cet intervalle est non convexe \fg applicable aux intervalles représentant une action qui se répète avec au moins deux occurrences disjointes, on serait capable d'exprimer ce que nous appellerons des \og actions répétées sporadique \fg,  comme dans l'exemple ci-haut, \og simmer 2--3 hours, stirring occasionally \fg : \og `simmer' \{contains\} `stir' \fg, où `stir' est un intervalle non convexe. Il pourrait en revanche être plus simple pour le raisonnement de représenter une action répétée sporadiquement sur un intervalle convexe spécialement qualifié : \og `simmer' \{di\} `stir (sporadiquement)' \fg.

\cite{bo99jnmr} ont pour leur part proposé une extension permettant d'exprimer des contraintes sur des intervalles cycliques, qui sont représentés sur un cercle plutôt que sur une droite. Ces c\_intervalles seraient utiles pour la représentation des actions répétées : par exemple, on pourrait représenter \og add milk and flour in alternance a little at a time \fg par \og `add milk' \{mmi\} `add flour' \fg, où \og mmi \fg représente la conjonction de \og m \fg et \og mi \fg. En revanche, les c\_intervalles ne permettent pas de représenter les relations de précédence temporelle.

\subsubsection{Contraintes métriques}

\cite{dmp91ai} ont proposé un formalisme, l'algèbre de distance, permettant de représenter et de raisonner sur des contraintes métriques exprimées sur des points, qui sont des variables réelles. Une contrainte binaire dans cette algèbre est une disjonction d'expressions sous la forme \og min $\leqslant X_j - X_i \leqslant$ max \fg. On pourrait donc représenter la durée d'une action comme \og simmer 2--3 hours \fg par \og $2 \leqslant I^+_{\text{simmer}} - I^-_{\text{simmer}} \leqslant 3$ \fg. Si toutes les contraintes sur les intervalles ne peuvent pas être exprimées (par exemple, la disjonction d'intervalles \{b,bi\}), il semble que les contraintes utiles à la représentation des recettes de cuisine soient toutes exprimables, bien que pas toujours de façon très intuitive. Cependant, il existe un formalisme simple proposé par \cite{kl91aaai} qui permet de représenter à la fois des contraintes métriques et des contraintes qualitatives sur des intervalles.

\subsection{Représentations proposées par la linguistique informatique}

Plusieurs formalismes temporels ont été proposés par des linguistes informaticiens spécifiquement pour représenter des informations temporelles extraites de textes, ou annoter des textes --- c'est-à-dire ajouter une couche d'information formelle. Le langage d'annotation le plus expressif à notre connaissance est TimeML \citep{pustejovsky04lot}. Celui-ci permet d'exprimer des relations entre des actions, des états, des intervalles, des points dans le temps et même des intervalles non convexes exprimés sous la forme d'\og ensembles de repères temporels \fg. En termes de relations pouvant être encodées, TimeML est sans doute comparable aux algèbres combinant contraintes qualitatives et quantitatives, mais il n'est pas associé à une sémantique ni à des règles d'inférences, bien que certaines applications existantes proposent de telles règles, notamment le système de question-réponse de \cite{hb05itqa}.

Voici par exemple les deux premières phrases de la recette de la figure~\ref{fig:recette} annotées en TimeML. Les balises EVENT soulignent ici les verbes, qui correspondent à des intervalles, alors que la balise SIGNAL souligne le mot exprimant la relation temporelle existant entre les événements correspondant à chacun des verbes, qui est formalisée par la balise TLINK --- la relation IS\_INCLUDED correspond exactement à la relation \{di\} en algèbre des intervalles.

$<$\texttt{EVENT eid="e1" class="OCCURENCE"}$>$ Brown $<$\texttt{/EVENT}$>$
\\$<$\texttt{MAKEINSTANCE eiid="ei1" eventID="e1" tense="INFINITIVE"\\ aspect="NONE" pos="VERB"/}$>$
hamburger and sausage with onion, garlic,\\ and all seasonings.
$<$\texttt{SIGNAL sid="s1"}$>$ Meanwhile $<$\texttt{/SIGNAL}$>$,
$<$\texttt{EVENT\\ eid="e2" class="OCCURENCE"}$>$ prepare $<$\texttt{/EVENT}$>$
$<$\texttt{MAKEINSTANCE\\ eiid="ei2" eventID="e2" tense="INFINITIVE" aspect="NONE"\\ pos="VERB"/}$>$
the pasta per pkg instructions.
$<$\texttt{TLINK\\ eventInstanceID="ei2" signalID="s1" relatedToEvent="ei1"\\ relType="IS\_INCLUDED"/}$>$

Les langages d'annotation permettent d'associer des contraintes directement à des mots. Pour raisonner sur des textes, et en particulier pour l'adaptation, cela constitue un avantage considérable : ainsi, toute modification dans le réseau de contrainte peut plus facilement être reportée sur le texte.

\subsection{Flux opérationnels \textit{(workflows)}}

\begin{figure}
	\includegraphics[width=\textwidth]{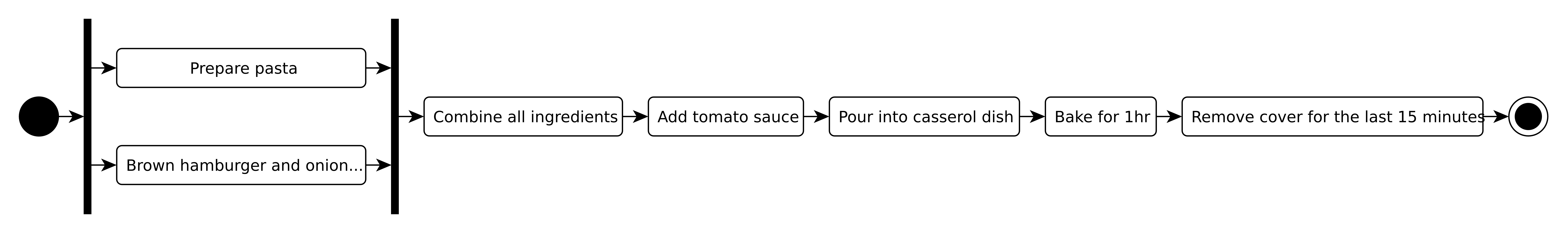}
	\caption{La recette de la figure~\ref{fig:recette} formalisée sous forme de flux opérationnel. Les actions entre les deux bandes noires sont conjointes, définissant un ordre partiel.}
	\label{fig:flux}
\end{figure}

Un flux opérationnel est une modélisation d'un ensemble d'actions qui doivent être accomplies, éventuellement par différents acteurs, pour réaliser une tâche. Si les possibilités expressives temporelles des flux sont assez restreintes, étant essentiellement limitées à spécifier l'ordre des actions ou éventuellement la possibilité de les accomplir en parallèle, ils présentent par ailleurs une plus grande richesse en terme de structures de contrôle. En particulier, il est facile d'exprimer des disjonctions et des boucles, qui sont les phénomènes qui nous posent le plus de problème pour la représentation algébrique temporelle des recettes de cuisine. Les flux opérationnels sont d'ailleurs déjà employés par \cite{minor10ccc} pour l'adaptation de recettes. Cependant, tout comme les formalismes d'annotation de texte, les flux ne sont pas à proprement parler associés à une sémantique et à des processus d'inférence. Il faut noter par ailleurs qu'il existe de nombreux langages de flux proposant différents niveaux d'expressivité, et que les langages les plus expressifs peuvent s'apparenter à des langages algorithmiques et donc poser des problèmes de décidabilité. La figure~\ref{fig:flux} montre le texte de recette de la figure~\ref{fig:recette} formalisé sous forme de flux opérationnel.

\section{Synthèse}
\label{sec:synth}

La table~\ref{tab:synth} reprend les principaux phénomènes à représenter dans les recettes identifiés à la section~\ref{sec:phen} afin de noter ceux pour lesquels nous croyons avoir identifié une manière de les exprimer dans les différents formalismes présentés dans la section~\ref{sec:form}. Bien entendu, le fait que nous n'ayons pas identifié une manière de représenter un phénomène donné dans un formalisme ne constitue pas une preuve que cette manière n'existe pas.

\begin{table}
\vspace{15mm}
\begin{tabular}{llllllll}
	& \begin{rotate}{45} Allen \end{rotate}
	& \begin{rotate}{45} $\mathcal{INDU}$ \end{rotate}
	& \begin{rotate}{45} Intervalles non convexes \end{rotate}
	& \begin{rotate}{45} Intervalles cycliques \end{rotate}
	& \begin{rotate}{45} Algèbre de distance \end{rotate}
	& \begin{rotate}{45} TimeML \end{rotate}
	& \begin{rotate}{45} Flux opérationnels \end{rotate}
	\\ \hline

	Durée qualitative & \checkmark & \checkmark & \checkmark & \checkmark  & \checkmark & \checkmark & \checkmark$^{\text{a}}$ \\ \hline
	Durée quantitative précise &  & \checkmark &  & & \checkmark & \checkmark & \\ \hline
	Durée quantitative imprécise &  & \checkmark &  &  & \checkmark & & \\ \hline
	Ordre total & \checkmark & \checkmark & \checkmark & & \checkmark & \checkmark & \checkmark \\ \hline
	Ordre partiel & \checkmark & \checkmark & \checkmark & & \checkmark & \checkmark & \checkmark \\ \hline
	Simultanéité  & \checkmark & \checkmark & \checkmark & & \checkmark & \checkmark &  \\ \hline\noalign{\smallskip}
	\parbox{13em}{Action répétée un nombre\\indéterminé de fois} & & & \checkmark & \checkmark & & \checkmark & \checkmark \\ \noalign{\smallskip}\hline
	Actions répétées en alternance & & & \checkmark & \checkmark & & & \checkmark \\ \hline
	Action répétée sporadiquement & & & \checkmark & & & \checkmark & \checkmark$^{\text{a}}$ \\ \hline
	Disjonction exclusive & & & & & &  & \checkmark \\ \hline
\end{tabular}
\caption{Adéquation entre les phénomènes temporels à représenter et les formalismes. Les types de durée sont définis à la section~\ref{subsec:duree}, les types d'ordre et la simultanéité à la section~\ref{subsec:ordre},  les types de répétition d'action à la section~\ref{subsec:repet} et la disjonction à la section~\ref{subsec:disj}.}
\caption*{\footnotesize a. On peut représenter ces phénomènes en combinant une boucle et une instruction de type \og no operation \fg.}
\label{tab:synth}

\end{table}

\section{Conclusion}
\label{sec:concl}

Notre première constatation est que chaque phénomène que nous souhaitons représenter est exprimable dans au moins un formalisme étudié, mais qu'aucun formalisme ne semble capable d'exprimer tous les phénomènes. La question qui se pose donc est de savoir s'il faut créer un nouveau formalisme ou étendre un formalisme existant, ou encore s'il est possible de combiner certains formalismes que nous avons étudiés. En particulier, \cite{ckms09cosit} se sont penchés sur le problème de la fusion de connaissances spatio-temporelles à travers l'utilisation de réseaux de contraintes qualitatives issues de différents formalismes. Ces travaux constituent une base intéressante sur laquelle nous comptons nous appuyer pour nos travaux futurs sur l'adaptation spatio-temporelle mais, plus spécifiquement, ils créent également un cadre permettant de combiner différentes algèbres qualitatives. À long terme, il serait intéressant d'étudier certains des formalismes que nous avons laissés de côté, par exemple ceux liés à la représentation des actions et des changements et ceux liés à la planification --- en particulier, il existe une littérature assez étendue sur la planification à partir de cas, qui pourrait constituer une approche complémentaire à celle que nous tentons de développer. Par exemple, le calcul des situations ou un formalisme apparenté pourrait être employé pour vérifier que la cohérence d'une recette est maintenue après son adaptation.

Nous avons concentré notre attention sur les recettes de cuisine, mais notre étude aurait pu s'appliquer à plusieurs autres types de textes, voire de cas non textuels. En effet, plusieurs ensembles de cas de type procédural, que ce soient des instructions d'assemblage, de préparation pharmaceutique ou de réparation de locomotive, pourraient bénificier d'une procédure d'adaptation automatique lorsque l'une des pré-conditions requises à l'exécution de la procédure n'est pas respectée. Les mêmes types de représentation et de raisonnement pourraient alors être employés. Par ailleurs, l'approche que nous souhaitons développer sera applicable également au raisonnement à partir de cas spaciaux ou spatio-temporels --- par exemple pour l'adaptation d'itinéraires ou l'aide à la prise de décision sur des projets d'aménagement agricole ---, soulevant des problèmes de représentation similaires.

\bibliography{biblio}
\end{document}